\RequirePackage{fix-cm}
\documentclass[smallextended]{svjour3}       
\smartqed  
\usepackage{cite}
\usepackage{array}
\usepackage{algorithmic}
\usepackage{amsmath,amssymb,amsfonts}
\usepackage{algorithm}
\usepackage{graphicx}
\usepackage{epstopdf}
\usepackage{subfigure}
\usepackage{multirow}
\usepackage{times}
\usepackage{color}
\usepackage{epsfig}
\begin{document}

\title{SA-IGA: A Multiagent Reinforcement Learning Method Towards Socially Optimal Outcomes}


\author{Chengwei~Zhang  \and Xiaohong~Li \and Jianye~Hao \and Siqi~Chen \and Karl~Tuyls \and Wanli Xue
}


\institute{Chengwei~Zhang, Xiaohong~Li and Wanli Xue\at School of Computer Science and Technology, Tianjin University \\
              \email{chenvy,xiaohongli,xuewanli@tju.edu.cn}           
           \and
           Jianye~Hao, Corresponding author \at School of Computer Software, Tianjin University \\
              \email{jianye.hao@tju.edu.cn}           
           \and
           Siqi~Chen \at School of Computer and Information Science, the Southwest University \\
              \email{siqichen@swu.edu.cn}           
           \and
           Karl~Tuyls \at University of Liverpool \\
              \email{k.tuyls@liverpool.ac.uk}           
}

\date{Received: date / Accepted: date}

\maketitle

\begin{abstract}
In multiagent environments, the capability of learning is important for an agent to behave appropriately in face of unknown opponents and dynamic environment. From the system designer's perspective, it is desirable if the agents can learn to coordinate towards socially optimal outcomes, while also avoiding being exploited by selfish opponents. To this end, we propose a novel gradient ascent based algorithm (SA-IGA) which augments the basic gradient-ascent algorithm by incorporating social awareness into the policy update process. We theoretically analyze the learning dynamics of SA-IGA using dynamical system theory and SA-IGA is shown to have linear dynamics for a wide range of games including symmetric games. The learning dynamics of two representative games (the prisoner's dilemma game and the coordination game) are analyzed in details. Based on the idea of SA-IGA, we further propose a practical multiagent learning algorithm, called SA-PGA, based on Q-learning update rule. Simulation results show that SA-PGA agent can achieve higher social welfare than previous social-optimality oriented Conditional Joint Action Learner (CJAL) and also is robust against individually rational opponents by reaching Nash equilibrium solutions.
\keywords{Multiagent Reinforcement Learning \and Social Welfare \and Gradient Ascent \and Nonlinear Analysis}
\end{abstract}

\section{Introduction}\label{sec-introduction}

In multiagent systems, the ability of learning is important for an agent to adaptively adjust its behaviors in response to coexisting agents and unknown environments in order to optimize its performance. Multiagent learning algorithms have received extensive investigation in the literature, and lots of learning strategies \cite{busoniu2008comprehensive,matignon2012independent,bloembergen2015evolutionary,Zhang2017FMRQ} have been proposed to facilitate coordination among agents.

The multi-agent learning criteria proposed in \cite{WOLF-PHC} require that an agent should be able to converge to a stationary policy against some class of opponents (\emph{convergence}) and the best-response policy against any stationary opponent (\emph{rationality}). If both agents adopt a rational learning strategy in the context of repeated games and also their strategies converge, then they will converge to a Nash equilibrium of the stage game. Indeed, convergence to Nash equilibrium has been the most commonly accepted goal to pursue in multiagent learning literature. Until now, a number of gradient-ascent based multiagent learning algorithms \cite{singh2000nash,WOLF-PHC,Abdallah2008MRL,zhang2010multi,} have been sequentially proposed towards converging to Nash equilibrium with improved convergence performance and more relaxed assumptions (less information is required). Under the same direction, another well-studied family of multiagent learning strategies is based on reinforcement learning (e.g., Q-learning \cite{Q-learning}). Representative examples include distributed Q-learning in cooperative games \cite{lauerRiedmiller}, minimax Q-learning in zero-sum games \cite{minmaxQ}, Nash Q-learning in general-sum games \cite{hu2003nash}, and other extensions \cite{littman2001friend,busoniu2008comprehensive}, to name just a few.

\begin{table}[h]
\centering
\begin{tabular}{ccccc}
\hline
\rule{0pt}{12pt}
\multirow{2}{0.6 in}{1's payoff \\ 2's payoff} &  & \multicolumn{2}{c}{Agent 2's actions} \\
\cline{3-4}
\rule{0pt}{12pt}
 & & C & D \\
\hline
\multirow{2}{0.5 in}{\\Agent 1's \\ actions}\\
 & C & 3/3 & 0/5 \\
\cline{2-4}
\rule{0pt}{12pt}
 & D & 5/0 & 1/1 \\
\cline{2-4}
\hline
\end{tabular}
\caption{The Prisoner's Dilemma Game} \label{tab:pdg1}
\end{table}

All the aforementioned learning strategies pursue converging to Nash equilibrium under self-play, however, Nash equilibrium solution may be undesirable in many scenarios. One well-known example is the prisoner's dilemma (PD) game shown in Table \ref{tab:pdg1}. By converging to the Nash equilibrium $(D, D)$, both agents obtain the payoff of 1, while they could have obtained a much higher payoff of 3 by coordinating on the non-equilibrium outcome $(C, C)$. In situations like the PD game, converging to the socially optimal outcome, i.e., the maximal total reward of all players, under self-play would be more preferred. To address this issue, one natural modification for a gradient-ascent learner is to update its policy along the direction of maximizing the sum of all agents' expected payoff instead of its own. However, in an open environment, the agents are usually designed by different parties and may have not the incentive to follow the strategy we design. The above way of updating strategy would be easily exploited and taken advantage by (equilibrium-driven) self-interested agents. Thus it would be highly desirable if an agent can converge to socially optimal outcomes under self-play and Nash equilibrium against self-interested agents to avoid being exploited.

In this paper, we propose a new gradient-ascent based algorithm (SA-IGA) which augments the basic gradient ascent algorithm by incorporating ``social awareness'' into the policy update process. Social awareness means that agents try to optimize social outcomes as well as its own outcome. A SA-IGA agent holds a social attitude to reflect its socially-aware degree, which can be adjusted adaptively based on the relative performance between its own and its opponent. A SA-IGA agent seeks to update its policy in the direction of increasing its overall payoff which is defined as the average of its individual and the social payoff weighted by its socially-aware degree. We theoretically show that for a wide range of games (e.g., symmetric games), the dynamics of SA-IGAs under self-play exhibits linear characteristics. For general-sum games, it may exhibit non-linear dynamics which can still be analyzed numerically. The learning dynamics of two representative games (the prisoner's dilemma game and the coordination game representing symmetric games and asymmetric games, respectively) are analyzed in details. Like previous theoretical multiagent learning algorithms, SA-IGA also requires additional assumption of knowing the opponent's policy and the game structure.

To relax the above assumption, we then propose a practical gradient ascent based multiagent learning strategy, called Socially-aware Policy Gradient Ascent (SA-PGA). SA-PGA relaxes the above assumptions by estimating the performance of its own and the opponent using Q-learning techniques. We empirically evaluate its performance in different types of benchmark games and simulation results show that SA-PGA agent outperforms previous learning strategies in terms of maximizing the social welfare and Nash product of the agents. Besides, SA-PGA is also shown to be robust against individually rational opponents and converges to Nash equilibrium solutions.

The remainder of the paper is organized as follows. Section \ref{relatedworks} generally reviews some related works about Gradient Ascent Reinforcement Learning algorithms. Section \ref{background} reviews normal-form game and the basic gradient ascent approach. Section \ref{sec-model} introduces the SA-IGA algorithm and analyzes its learning dynamics theoretically. Section \ref{sec-algorithm} presents the practical multiagent learning algorithm SA-PGA in details. In Section \ref{experiment}, we extensively evaluate the performance of SA-PGA under various benchmark games. Lastly we conclude the paper and point out future directions in Section \ref{conclusion}.

\section{Related Works}\label{relatedworks}

The first gradient ascent multiagent reinforcement learning algorithm is Infinitesimal Gradient Ascent (IGA\cite{singh2000nash}), in which each learner updates its policy towards the gradient direction of its expected payoff. The purpose of IGA is to promote agents to converge to a particular Nash Equilibrium in a two-player two-action normal-form game. IGA has been proved that agents will converge to Nash equilibrium or if the strategies themselves do not converge, then their average payoffs will nevertheless converge to the average payoffs of Nash equilibrium. Soon after, M. Zinkevich et al. \cite{Zinkevich2003Online} propose an algorithm called Generalized Infinitesimal Gradient Ascent(GIGA), which extends IGA to the game with an arbitrary number of actions.

Both IGA and GIGA can be combined with the Win or Learn Fast (WoLF) heuristic in order to improve performance in stochastic games (Wolf-IGA\cite{WOLF-PHC}, Wolf-GIGA\cite{Bowling2004Convergence}). The intuition behind WoLF principle is that an agent should adapt quickly when it performs worse than expected, whereas it should maintain the current strategy when it receives payoff better than the expected one. By altering the learning rate according to the WoLF principle, a rational algorithm can be made convergent.
The shortage of WoLF-IGA or WoLF-GIGA is that these two algorithms require a reference policy, i.e., they require the estimation of Nash equilibrium strategies and corresponding payoffs.
To this end, Banerjee et al\cite{Banerjee2003Adaptive} propose an alternative criterion of WoLF-IGA, named Policy Dynamics based WoLF(PDWoLF), that can be accurately computed and guarantees convergence. The Weighted Policy Learner (WPL\cite{Abdallah2008MRL}) is another variation of IGA that also modulates the learning rate, meanwhile, it does not require a reference policy. Both of the WoLF and WPL are designed to guarantee convergence in stochastic repeated games.

Another direction for extending IGA is making improvements from the learning value functions. Zhang et al\cite{zhang2010multi} propose a gradient-based learning algorithm by adjusting the expected payoff function of IGA, named Gradient Ascent with Policy Prediction Algorithm(IGA-PP). The algorithm is designed for games with two agents. The key idea behind this algorithm is that a player adjusts its strategy in response to forecasted strategies of the other player, instead of its current ones. It has been proved that, in two-player, two-action, general-sum matrix games, IGA-PP in self-play or against IGA would lead players' strategies to converge to a Nash equilibrium. Like other MARL algorithms, besides the common assumption, this algorithm also has additional requirements that a player knows the other player¡¯s strategy and current strategy gradient (or payoff matrix) so that it can forecast the other player¡¯s strategy.

All the aforementioned learning strategies pursue converging to Nash equilibriums. In contrast, in this work, we seek to incorporate the social awareness into GA-based strategy update and aim at improving the social welfare of the players under self-play rather than pursuing Nash equilibrium solutions. Meanwhile, individually rational behavior is employed when playing against a selfish agent. Similar idea of adaptively behaving differently against different opponents was also employed in previous algorithms \cite{littman2001friend,conitzer2007awesome,powers2005learning,chakraborty2014multiagent}. However, all the existing works focus on maximizing an agent's individual payoff against different opponents in different types of games, but do not directly take into consideration the goal of maximizing social welfare (e.g., cooperate in the prisoner's dilemma game).

\section{Background}\label{background}

In this section we introduce the necessary background for our contribution. First, we gave an overview of the relevant game theory definition. Then a brief review of gradient ascent based MARL (GA-MARL) algorithm is given.

\subsection{Game theory}

Game theory provides a framework for modeling agents' interaction, which was used by previous researchers in order to analyze the convergence properties of MARL algorithms 
\cite{singh2000nash,WOLF-PHC,Abdallah2008MRL,zhang2010multi}. A game specifies, in a compact and simple manner, how the payoff of an agent depends on other agents¡¯ actions. A (normal form) game is defined by the tuple $ < N,{A_1},...,{A_N},{R_1},...,{R_N} > $, where $N$ is the number of players in the game, $A_i$ is the set of actions available to agent $i$, and ${R_i}:{A_1} \times ... \times {A_N} \to \mathbb{R} $ is the reward (payoff) of agent $i$ which is defined as a function of the joint action executed by all agents. If the game has only two agents, then it is convenient to define their reward functions as a payoff matrix as follows,
\[R_{i}=\{ r_i^{jk} \}_{{|A_1|} \times {|A_2|}}\]
where $i\in \{1,2\}$, $j\in A_j$ and $k\in A_k$. Each element $r_i^{jk}$ in the matrix represents the payoff received by agent $i$, if agent $i$ plays action $j$ and its opponent plays action $k$.

A $policy$ (or a $strategy$) of an agent $i$ is denoted by $\pi_i:A_i \to \left[0,1\right]$, which maps its actions to a probability. The probability of choosing an action $k$ according to policy $\pi_i$ is $\pi_i(k)$. A policy is deterministic or pure if the probability of playing one action is $1$ while the probability of playing other actions is 0, (i.e. $\exists \pi_i(k)= 1$ AND $\forall l\ne k, \pi_i(l)= 0$), otherwise the policy is stochastic or mixed. The joint policy of all agents is the collection of individual agents' policies, which is defined as $\pi=<\pi_1,...,\pi_N>$. For continence, the joint policy is usually expressed as $\pi=<\pi_i,\pi_{-i}>$, where $\pi_{-i}$ is the collection of all policies of agents other than agent $i$.

The $expected$ $payoff$ of an agent is defined as the reward averaged over the joint policy. Let $A_{-i} = \{<a_1,...,a_N>: a_j \in A_j \wedge  i \ne j\}$, if agents follow a joint policy $\pi$, then the $expected$ $payoff$ of agent $i$ would be,
${V_i}\left( \pi  \right) = \sum\nolimits_{{a_i} \in {A_i}} {\sum\nolimits_{{a_{ - i}} \in {A_{ - i}}} {{\pi _i}\left( {{a_i}} \right){\pi _{ - i}}\left( {{a_{ - i}}} \right){R_i}\left( {{a_i},{a_{ - i}}} \right)} } $, where ${R_i}\left( {{a_i},{a_{ - i}}} \right)=r_i^{a_ia_{ - i}}$.

The goal of each agent is to find such a policy that maximizes the player¡¯s expected payoff. Ideally, we want all agents to reach the equilibrium that maximizes their individual payoffs. However, when agents do not communicate and/or agents are not cooperative, reaching a globally optimal equilibrium is not always attainable. An alternative goal is converging to the Nash Equilibrium (NE), which is by definition a local maximum across agents. A joint strategy is called a $Nash$ $Equilibrium$ (NE), if no player can get a better expected payoff by changing its current strategy unilaterally. Formally,
$\pi^{*}=\left ( \pi_{i}^{*},\pi_{-i}^{*} \right )$ is a NE, iff $\forall i \forall \pi_{i}$: $V_{i}\left (\pi_{i}^{*},\pi_{-i}^{*} \right )\geq V_{i}\left (\pi_{i},\pi_{-i}^{*} \right)$ . An NE is pure if all its constituting policies are pure. Otherwise the NE is called mixed or stochastic. Any game has at least one Nash equilibrium, but may not have any pure equilibrium.

Next subsection, we introduce the Gradient Ascent based MARL algorithm (GA-MARL), together with a brief review of the dynamic analysis of GA-MARL.
\subsection{Gradient Ascent (GA) MARL Algorithms}

Gradient ascent MARL algorithms (GA-MARL) learn a stochastic policy by directly following the expected reward gradient. The ability to learn a stochastic policy is particularly important when the world is not fully observable or has a competitive nature. The basic GA-MARL algorithm whose dynamics were analyzed is the Infinitesimal Gradient Ascent(IGA \cite{singh2000nash}) . When a game is repeatedly played, an IGA player updates its strategy towards maximizing its expected payoffs. A player $i$ employing GA-based algorithms updates its policy towards the direction of its expected reward gradient, as illustrated by the following equations,

\begin{equation}
\label{eq2.1deltaPi}
\Delta \pi_{i}^{\left(t+1\right)}\leftarrow \alpha \frac{\partial V_{i}\left ( \pi^{\left(t\right)}\right )}{\partial \pi_{i}}
\end{equation}
\begin{equation}
\label{eq2.2PiUpdate}
 \pi_{i}^{\left(t+1\right)}\leftarrow \Pi_{\left [0,1\right ]}\left ( \pi_{i}^{\left(t\right)}+\Delta \pi_{i}^{\left(t+1\right)}\right)
\end{equation}
where parameter $\alpha $ is the gradient step size, and $\Pi_{[0,1]}$ is the projection function mapping the input value to the valid probability range of $[0,1]$, used to prevent the gradient moving the strategy out of the valid probability space. Formally, we have,
\begin{equation}
\Pi_{\left [0,1 \right]}\left(x\right)=argmin_{z\in\left[0,1\right]}\left|x-z\right|
\end{equation}

Singh, Kearns, and Mansour \cite{singh2000nash} examined the dynamics of using gradient ascent in two-player, two-action, iterated matrix games. We can represent this problem as two matrices,
\[{R_i} = \left[ {\begin{array}{*{20}{c}}
{r_i^{11}}&{r_i^{12}}\\
{r_i^{21}}&{r_i^{22}}
\end{array}} \right],i \in \{ 1,2\} \]

We refer to the joint policy of the two players at time $t$ by the probabilities of choosing the first action $\left(p_1^t, p_2^t\right)$, where $\pi_i = \left(p_i^t, 1-p_i^t\right)$, $i \in \{1,2\}$ is the policy of player $i$. The $t$ notation will be omitted when it does not affect clarity (for example, when we are considering only one point in time). Then, for the two-player two-action case, the above way of GA-based updating in Equations \ref{eq2.1deltaPi} and \ref{eq2.2PiUpdate} can be simplified as follows,
\begin{equation}
\label{eq2.3PiUpdateNew}
p_{i}^{\left(t+1\right)}\leftarrow \Pi_{\left [0,1\right]}\left(p_{i}^{\left(t\right)}+\alpha\left(u_{i}p_{-i}^{\left(t\right)}+c_{i}\right)\right)
\end{equation}
where $u_{i}=r_{i}^{11}+r_{i}^{22}-r_{i}^{12}-r_{i}^{21}$, $c_{i}=r_{i}^{12}-r_{i}^{22}$.

In the case of infinitesimal gradient step size ($\eta \rightarrow 0$), the learning dynamics of the players can be modeled as a system of differential equations, i.e. $\dot{p}_{i}=u_{i}p_{-i}+c_{i}$, $i\in \{1,2\}$, which can be analyzed using dynamic system theory \cite{Coddington1955Theory}. It is proved that the agents will converge to a Nash equilibrium, or if the strategies themselves do not converge, then their average payoffs will nevertheless converge to the average payoffs of a Nash equilibrium \cite{singh2000nash}.

Combined with Q-learning\cite{Watkins1989Learning}, researchers propose a practical learning algorithm, i.e. the policy hill-climbing algorithm (PHC)\cite{WOLF-PHC}, which is a simple extension of IGA and is shown in Table \ref{alg:PHC}.

\begin{algorithm}
\caption{PHC for player $i$}
\label{alg:PHC}
\begin{algorithmic}[1]
\STATE  Lets $\alpha,\beta \in \left ( 0,1 \right )$ be learning rates.
\STATE Initialize,\\
 $Q_{i}\left (a\right )\leftarrow 0$, $\pi_i(a)\leftarrow \frac{1}{|A_i|}$.
\REPEAT
\STATE Select action $a\in A_i$ according to mixed strategy $\pi_i$ with suitable exploration.
\STATE Observing reward $r$. Update $Q$,\\
  $Q_{i}\left ( a \right ) \leftarrow \left (1-\beta \right )Q_{i}\left ( a \right )+\beta r$.\\
\STATE Update $\pi_i$ according to gradient ascent strategy,\\
  $\pi_i \left ( a \right ) \leftarrow \Pi_{\left[0,1\right]}[\pi_i \left ( a \right )- \alpha]$, if $a \ne \mathop{argmax}\limits _{a'\in A}Q\left( {a'} \right)$,\\
  $\pi_i \left ( a \right ) \leftarrow 1-\sum\limits_{a' \ne a} {\pi \left( {a'} \right)} $, if $a = \mathop{argmax} \limits_{a'\in A}Q\left( {a'} \right)$.\\
\UNTIL{the repeated game ends}
\end{algorithmic}
\end{algorithm}

The algorithm performs hill-climbing in the space of mixed policies, which is similar to gradient ascent, but does not require as much knowledge. Q values are maintained just as in normal Q-learning. In addition the algorithm maintains the current mixed policy. The policy is improved by increasing the probability that it selects the highest valued action according to a learning rate $\alpha \in \left(0, 1\right]$. After that, the policy is mapped back to the valid probability space. This technique, like Q-learning, is rational and will converge to an optimal policy if other players are playing stationary strategies. The algorithm guarantees the $Q$ values will converge to $Q^{*}$ (the local optimal value of $Q$) with a suitable exploration policy. $\pi$ will converge to a policy that is greedy according to $Q$, which is converging to $Q^{*}$, and therefore will converge to a best response. PHC is rational and has no limit on the number of agents and actions.

\section{Socially-aware Infinitesimal Gradient Ascent (SA-IGA)}
\label{sec-model}
\label{SA-IGA}
In our daily life, people usually do not always behave as a purely individually rational entity and seek to achieve Nash equilibrium solutions. For example, when two person subjects play a PD game, reaching mutual cooperation may be observed frequently. Similar phenomena have also been observed in extensive human-subject based experiments in games such as the Public Goods game\cite{Hauert2003Prisone} and Ultimatum game\cite{Alvard2004The}, in which human subjects are usually found to obtain much higher payoff by mutual cooperation rather than pursuing Nash equilibrium solutions. If the above phenomenon is transformed into computational models, it indicates that an agent may not only update its policy in the direction of maximizing its own payoff, but also take into consideration other's payoff. We call this type of agents as socially-aware agents.

In this paper, we incorporate the social awareness into the gradient-ascent based learning algorithm. In this way, apart from learning to maximizing its individual payoff, an agent is also equipped with the social awareness so that it can (1) reach mutually cooperative solutions faced with other socially-aware agents (self-play); (2) behave in a purely individually rational manner when others are purely rational.

Specifically, for each SA-IGA agent $i$, it distinguishes two types of expected payoffs, namely $V_{i}^{\mathrm{idv}}$ and $V_i^{\mathrm{soc}}$. Payoffs $V_{i}^{\mathrm{idv}}\left(\pi\right)$ and $V_i^{\mathrm{soc}}\left(\pi\right)$ represent the individual and social payoff (the average payoff of all agents) that agent $i$ perceives under the joint strategy $\pi$ respectively. The payoff $V_{i}^{\mathrm{idv}}\left(\pi\right)$ follows the same definition as IGA and the payoff $V_i^{\mathrm{soc}}\left(\pi\right)$ is defined as the average of the individual payoffs of all agents.
\begin{equation}
\label{eq3.1Vsoc}
V_i^{\mathrm{soc}}\left(\pi\right)=\frac{1}{N}\sum\limits_i {{V_i^{\mathrm{idv}}}\left( \pi  \right)},
\end{equation}

Each agent $i$ adopts a social attitude $w_i \in [0,1]$ to reflect its socially-aware degree. The social attitude intuitively models an agent's social friendliness degree towards others. Specifically, it is used as the weighting factor to adjust the relative importance between $V_{i}^{\mathrm{idv}}$ and $V_i^{\mathrm{soc}}$, and agent $i$'s overall expected payoff is defined as follows,
\begin{equation}
\label{eq3.2Vi}
V\left(\pi\right)=\left(1-w_i\right)V_{i}^{\mathrm{idv}}\left(\pi\right)+w_iV_i^{\mathrm{soc}}\left(\pi\right)
\end{equation}

Each agent $i$ updates its strategy in the direction of maximizing the value of $V_{i}$. Formally we have,

\begin{eqnarray}
\label{eq3.3strategy}
\begin{split}
&\Delta \pi_{i}\leftarrow \alpha_{\pi} \frac{\partial V_{i}\left(\pi\right )}{\partial \pi_{i}}, \pi_{i}\leftarrow \Pi_{\left [0,1\right ]}\left ( \pi_{i}+\Delta \pi_{i}\right)
\end{split}
\end{eqnarray}
where parameter $\alpha_{\pi}$ is the gradient step size of $\pi_{i}$. If $w_i= 0$, it means that the agent seeks to maximize its individual payoff only, which is reduced to the case of traditional gradient-ascent updating; if $w_i= 1$, it means that the agent seeks to maximize the sum of the payoffs of both players.

Finally, each agent $i$'s socially-aware degree is adaptively adjusted in response to the relative value of $V_i^{\mathrm{idv}}$ and $V_i^{\mathrm{soc}}$ as follows. During each round, if player $i$'s own expected payoff $V_{i}^{\mathrm{idv}}$ exceeds the value of $V_i^{\mathrm{soc}}$, then player $i$ increases its social attitude $w_{i}$, (i.e., it becomes more social-friendly because it perceives itself to be earning more than the average). Conversely, if $V_{i}^{\mathrm{idv}}$ is less than $V_i^{\mathrm{soc}}$, then the agent tends to care more about its own interest by decreasing the value of $w_{i}$. Formally,
\begin{equation}
\label{eq3.4w-adjust}
w_i^{t + 1} \leftarrow \Pi_{\left [0,1\right ]}\left ( w_i^t + {\alpha _w}\left( {V_i^{idv} - V_i^{soc}} \right)\right)
\end{equation}
where parameter $\alpha_{w}$ is the learning rate of $w_{i}$.

\subsection{Theoretical Modeling and Analysis of SA-IGA}
An important aspect of understanding the behavior of a multiagent learning algorithm is theoretically modeling and analyzing its underlying dynamics \cite{tuyls2003selection,rodrigues2009dynamic,bloembergen2015evolutionary}. In this section, we first show that the learning dynamics of SA-IGA under self-play can be modeled as a system of differential equations. To simplify analysis, we only considered two-player-two-action games.

Based on the adjustment rules in Equation (\ref{eq3.3strategy}) and (\ref{eq3.4w-adjust}), the learning dynamics of a SA-IGA agent can be modeled as a set of equations in (\ref{eq3.5update}). For ease of exposition, we concentrate on an unconstrained update equations by removing the policy projection function which does not affect our qualitative analytical results. Any trajectory with linear (non-linear) characteristic without constraints is still linear (non-linear) when a boundary is enforced.
\begin{equation}
\label{eq3.5update}
\begin{split}
&\Delta \pi_{i}^{\left(t+1\right)}\leftarrow \alpha_{\pi} \frac{\partial V_{i}\left(\pi^{\left(t\right)}\right )}{\partial \pi_{i}}\\
&\Delta w_i^{t+1}\leftarrow \alpha_{w} (V_{i}^{\mathrm{idv}}-V_i^{\mathrm{soc}})\\
&\pi_{i}^{\left(t+1\right)}\leftarrow \pi_{i}^{\left(t\right)}+\Delta \pi_{i}^{\left(t+1\right)}\\
&w_{i}^{\left(t+1\right)}\leftarrow w_{i}^{\left(t\right)}+\Delta w_{i}^{\left(t+1\right)}
\end{split}
\end{equation}

Substituting $V_i^{\mathrm{idv}}$ and $V_i^{\mathrm{soc}}$ by their definitions (Equations \ref{eq2.3PiUpdateNew} and \ref{eq3.1Vsoc}), the learning dynamics of two SA-IGA agents can be expressed as follows,
\begin{equation}
\label{eq3.6update}
\begin{split}
&\Delta p_{i}^{t+1}=\alpha_{p} \cdot \left [\left(u_{i}+\frac{u_{-i}-u_{i}}{2}w_{i}^{t}\right )p_{-i}^{t}+\frac{d_{-i}-c_{i}}{2}w_{i}^{t}+c_{i}\right]\\
&\Delta w_{i}^{t+1}=\alpha_{w} \cdot \left[\left (u_{i}-u_{-i}\right)p_{i}^{t}p_{-i}^{t}+\left(c_{i}-d_{-i}\right)p_{i}^{t}+\left(c_{-i}-d_{i}\right)p_{-i}^{t}+e_i\right]\\
\end{split}
\end{equation}
where $u_{i}=r_{i}^{11}+r_{i}^{22}-r_{i}^{12}-r_{i}^{21}$, $c_{i}=r_{i}^{12}-r_{i}^{22}$,$d_{i}=r_{i}^{21}-r_{i}^{22}$, and $e_i=r_{i}^{22}-r_{-i}^{22}$ with $i\in\left\{1,2\right\}$.

As $\alpha_{\pi} \rightarrow 0$ and $\alpha_{w} \rightarrow 0$, it is straightforward to show that the above equations become differential. Thus the unconstrained dynamics of the strategy pair and social attitudes as a function of time is modeled by the following system of differential equations:
\begin{equation}
\label{eq3.7differential}
\begin{split}
&\dot{p}_{i}=\left(u_{i}+\frac{u_{-i}-u_{i}}{2}w_{i}\right )p_{-i}+\frac{d_{-i}-c_{i}}{2}w_{i}+c_{i}\\
&\dot{w}_{i}=\varepsilon \cdot \left[\left(u_{i}-u_{-i}\right)p_{i}p_{-i}+\left(c_{i}-d_{-i}\right)p_{i}+\left(c_{-i}-d_{i}\right)p_{-i}+e_i\right]\\
\end{split}
\end{equation}
where $\varepsilon=\frac{\alpha_{w}}{\alpha_{p}}>0$.

Based on the above theoretical modeling, next we analyze the learning dynamics of SA-IGA qualitatively as follows.
\begin{theorem}
SA-IGA has non-linear dynamics when $u_{1}\neq u_{2}$.
\begin{proof}
: From differential equations in (\ref{eq3.7differential}), it is straightforward to verify that the dynamics of SA-IGA learners are non-linear when $u_{1}\neq u_{2}$ due to the existence of $w_{1}p_{2}$, $w_{2}p_{1}$ and $p_{1}p_{2}$ in all equations.
\end{proof}
\end{theorem}

Since SA-IGA's dynamics are non-linear when $u_{1}\neq u_{2}$, in general we cannot obtain a closed-form solution, but we can still resort to solve the equations numerically to obtain useful insight of the system's dynamics. Moreover, a wide range of important games fall into the category of $u_{1} = u_{2}$, in which the system of equations become linear. Therefore, it allows us to use dynamic system theory to systematically analyze the underlying dynamics of SA-IGA.

\begin{theorem}
SA-IGA has linear dynamics when the game itself is symmetric.
\begin{proof}
: A two-player two-action symmetric game can be represented in Table \ref{tab:sygame} in general. It is obvious to check that it satisfies the constraint of $u_{1} = u_{2}$, given that $u_{i}=r_i^{11}+r_{i}^{22}-r_{i}^{12}-r_{i}^{21}$, $ i\in\{1,2\}$. Thus the theorem holds.
\end{proof}
\end{theorem}

\begin{table}
\centering
\begin{tabular}{ccccc}
\hline
\rule{0pt}{12pt}
\multirow{2}{0.6 in}{1's payoff \\ 2's payoff} &  & \multicolumn{2}{c}{Agent 2's actions} \\
\cline{3-4}
\rule{0pt}{12pt}
 & & action 1 & action 2 \\
\hline
\multirow{2}{0.5 in}{\\Agent 1's \\ actions}\\
 & action 1 & a/a & b/c \\
\cline{2-4}
\rule{0pt}{12pt}
 & action 2 & c/b & d/d \\
\cline{2-4}
\hline
\end{tabular}
\caption{The General Form of a Symmetric Game} \label{tab:sygame}
\end{table}

\subsection{Dynamics Analysis of SA-IGA}

Previous section mainly analyzed the dynamics of SA-IGA in a qualitative manner. In this section, we move to provide detailed analysis of SA-IGA's learning dynamics. We first summarize a generalized conclusion for symmetric games, and then analysis symmetric circumstances in two representative games: the Prisoner's Dilemma game and the Symmetric Coordination game. For asymmetric circumstances, because the complexity of nonlinear problem analysis, we only focus on the general coordination game (Table \ref{tab:codgame}). Specifically we analyze the SA-IGA's learning dynamics of those games by identifying the existing equilibrium points, which provides useful insights into understanding of SA-IGA's dynamics.


For symmetric games, we have the following conclusion,

\begin{theorem}
\label{Symmetricgame}
The dynamics of SA-IGA algorithm under self-play under a symmetric game have three types of equilibrium points:
\begin{enumerate}
\item $\left\{ {\left( {0,0,w_1^*,w_2^*} \right)\left| {\frac{{c - b}}{2}w_i^* + b - d < 0,w_i^* \in [0,1]} \right.}\right\}$; \\
      $\left\{ {\left( {1,1,w_1^*,w_2^*} \right)\left| {\frac{{c - b}}{2}w_i^* + a - c > 0,w_i^* \in [0,1]} \right.}\right\}$;
\item $\left\{ {\left( {1,0,0,1} \right),\left( {0,1,1,0} \right)} \right\}$, if $ c>b>d \wedge b+c>2a $;\\
      $\left\{ {\left( {1,0,1,0} \right),\left( {0,1,0,1} \right)} \right\}$, if $ b>c>a \wedge b+c>2d $;
\item $\left\{ {\left( {{p^*},{p^*},{w^*},{w^*}} \right)\left| {{p^*} = \frac{{b - c}}{{2u}}{w^*} + \frac{{d - b}}{u},{p^*},{w^*} \in [0,1]} \right.} \right\}$,
\end{enumerate}

where $u=a+d-b-c$. The first and second types of equilibrium points are stable, while the last is not. We say an equilibrium point is stable if once the strategy starts "close enough" to the equilibrium (within a distance $\delta$ from it), it will remain "close enough" to the equilibrium point forever.
\begin{proof}
: Following the system of differential equations in Equations (\ref{eq3.7differential}), we can express the dynamics of SA-IGA in Symmetric game as follows:
\begin{equation}
\label{weifen}
\begin{split}
&\dot{p}_{i}=up_{-i}+\frac{c-b}{2}w_{i}+b-d\\
&\dot{w}_{i}=\varepsilon \left(b-c\right)\left(p_{i}-p_{-i}\right)\\
\end{split}
\end{equation}
where $\varepsilon=\frac{\eta_{w}}{\eta_{p}} >0$,$u=a+d-b-c$, $i\in \{1,2\}$.

We start with proving the last type of equilibrium points: If there exist an equilibrium $eq=\left(p_{1}^{*},p_{2}^{*},w_{1}^{*},w_{2}^{*}\right)^{T}\in\left(0,1\right)^{4}$, then we have $\dot{p}_{i}\left(eq\right)=0$ and $\dot{w}_{i}\left(eq\right)=0$, $i\in \{1,2\}$. By solving the above equations, we have $p_{1}^{*}=p_{2}^{*}=\frac{b-c}{2u}w^{*}+\frac{d-b}{u}$ and $w^{*}=w_{1}^{*}=w_{2}^{*}$. Since $p_{1}^{*},p_{2}^{*}\in \left (0,1 \right )$, then we have,
\[
0<\frac{b-c}{2u}w^{*}+\frac{d-b}{u}<1
\]

Then $eq=\left(p_{1}^{*},p_{2}^{*},w_{1}^{*},w_{2}^{*}\right)^{T}$ is an equilibrium. The stability of $eq$ can be verified using theories of non-linear dynamics\cite{shilnikov2001methods}. By expressing the unconstrained update differential equations in the form of $\dot{x}=Ax+B$, we have
\[
A=\begin{bmatrix}
0&u&c-b&0 \\
u&0&0&c-b \\
\varepsilon\left(b-c\right)&\varepsilon\left(c-b\right)&0&0\\
\varepsilon\left(c-b\right)&\varepsilon\left(b-c\right)&0&0
\end{bmatrix}
\]
After calculating matrix $A$'s eigenvalue, then we have $\lambda_{1}=0$, $\lambda_{2}=u$, $\lambda_{3}=-\frac{u}{2}+k$ and $\lambda_{4}=-\frac{u}{2}-k$, where $k$ is a constant. Since there exist an eigenvalue $\lambda>0$, the equilibrium $eq$ is not stable.

Next we turn to consider cases that equilibriums are in the boundary. In these cases, we need to put the projection function back. If $p_{i}=0, i \in \left\{1,2\right\}$, according to the known conditions, we have $w\frac{{c - b}}{2}w_i^*<d-b$. Combined with the unconstrained update differential equations \ref{weifen}, we have $\lim_{t \to \infty } \dot{p_{i}}<0$, then $p_{i}$ remains unchanged. And because $p_{1}=p_{2}=0$, then for $\forall w_{i}\in \left [ 0,1 \right]$, $\dot{w}_{i}=0$, then $\left(\left(0,0,w_{1}^{*},w_{2}^{*}\right)\right)$ is an equilibrium.

Because $w\frac{{c - b}}{2}w_i^*<d-b$, there exist a $\delta>0$, and a set $U\left (eq,\delta  \right )= \left\{ x\in\left [ 0,1 \right ]^{4}| \right.$ $\left. \left | x-eq \right |<\delta \right\}$, that for $\forall x\in U\left (eq,\delta  \right )$, $\lim_{p_{i}} \dot{p_{i}}<0$. Thus $p$ will stabilize on the point of 0. Also, as $\lim_{t\rightarrow 0} \dot{w_{i}}=\left(b-c\right)\lim_{t\rightarrow 0} \left(p_{1}-p_{2}\right)=\left(b-c\right)\lim_{t\rightarrow 0} \left(0-0\right)=0$, $w$ also stable, and thus the equilibrium $eq$ is stable.

The case that $p_i=1,i\in \{1,2\}$ can be proved similarly, which is omitted here.

For the case $p_1=1 \wedge p_2=0$, if $\left(\left(1,0,w_{1}^{*},w_{2}^{*}\right)\right)$ is an equilibrium, combined with the unconstrained update differential equations \ref{weifen}, we have $\dot{w_{1}}=-\dot{w_{2}}$, which means that $w_i$ will keeps changing until $w_1=1 \wedge w_2=0$ or $w_1=0 \wedge w_2=1$. If $\left(\left(1,0,0,1\right)\right)$ is an equilibrium, then $\dot{p_{1}}>0 \wedge \dot{p_{2}}<0$ and $\dot{w_{1}}<0 \wedge \dot{w_{2}}>0$. Take into Equations \ref{weifen}, we get $c>b>d \wedge b+c>2a$. Other case are the same, thus we it omit here.

The stability of the second type of equilibriums can be proved by the way as the first type one, which is omitted here.
\end{proof}
\end{theorem}

From Theorem \ref{Symmetricgame}, we know that there are three types of equilibriums if both players play SA-IGA policy, while only the first and second types of equilibrium points are stable. Besides, all equilibriums of the first two types are pure strategies, i.e., the probability $p_i$ for selecting action 1 for agent $i\in \{1,2\}$ equals to $1$ or $0$. Notably, the range of $w$ (the social attitude) in these three types of equilibriums may be overlapped, resulting in that the final convergence of the algorithm also depends on the value of $p$. Next we concentrate on details of two representative symmetric games: the Prisoner's Dilemma (PD) game and the Symmetric Coordination game.

The Prisoner's Dilemma (PD) game is a symmetric game whose parameters meet the conditions: $c>a>d>b$. Combined with Theorem \ref{Symmetricgame}, we have the following conclusion,
\begin{corollary}
\label{pdgdynamics}
The dynamics of SA-IGA algorithm under Prisoner's Dilemma (PD) game have two types of stable equilibrium points:
\begin{enumerate}
\item $\left(0,0,w_{1}^{*},w_{2}^{*}\right)$, if $w_{1}^{*},w_{2}^{*}<min\left \{ \frac{2\left(c-a\right)}{c-b},\frac{2\left(d-b\right)}{c-b} \right \}$;
\item $\left(1,1,w_{1}^{*},w_{2}^{*}\right)$, if $w_{1}^{*},w_{2}^{*}>max\left \{ \frac{2\left(c-a\right)}{c-b},\frac{2\left(d-b\right)}{c-b} \right \}$;
\end{enumerate}
\begin{proof}
: Because the PD game is a symmetric game, we can use conclusions of Theorem \ref{Symmetricgame} directly. From Theorem \ref{Symmetricgame}, we can see that the PD game have two types of stable equilibrium points:

\begin{enumerate}
\item $\left\{ {\left( {0,0,w_1^*,w_2^*} \right)\left| {\frac{{c - b}}{2}w_i^* + b - d < 0,w_i^* \in [0,1]} \right.}\right\}$; \\
      $\left\{ {\left( {1,1,w_1^*,w_2^*} \right)\left| {\frac{{c - b}}{2}w_i^* + a - c > 0,w_i^* \in [0,1]} \right.}\right\}$;
\item $\left\{ {\left( {1,0,0,1} \right),\left( {0,1,1,0} \right)} \right\}$, if $ c>b>d \wedge b+c>2a $;\\
      $\left\{ {\left( {1,0,1,0} \right),\left( {0,1,0,1} \right)} \right\}$, if $ b>c>a \wedge b+c>2d $;
\end{enumerate}

For the first type of equilibrium, take $c>a>d>b$ into conditions in above formulas, we have: if $w_{1}^{*},w_{2}^{*}<min\left \{ \frac{2\left(c-a\right)}{c-b},\frac{2\left(d-b\right)}{c-b} \right \}$, then $\left(0,0,w_{1}^{*},w_{2}^{*}\right)$ is an stable equilibrium; else if $w_{1}^{*},w_{2}^{*}>max\left \{ \frac{2\left(c-a\right)}{c-b},\frac{2\left(d-b\right)}{c-b} \right \}$, then $\left(0,0,w_{1}^{*},w_{2}^{*}\right)$ is an stable equilibrium.

For the second type of equilibrium, take $c>a>d>b$ into consideration, we found that the conditions are in conflict with each other, which means there is no such type of equilibriums under Prisoner's Dilemma (PD) game.
\end{proof}
\end{corollary}

Intuitively, for a PD game, from Corollary \ref{pdgdynamics}, we know that if both SA-IGA players are initially sufficiently social-friendly (the value of w is large than a certain threshold), then they will always converge to mutual cooperation of $(C,C)$. In other words, given that the value of $w$ exceeds certain threshold, the strategy point of $(1,1)$ (or $(C,C)$) in the strategy space is asymptotically stable. If both players start with a low socially-aware degree ($w$ is smaller than certain threshold), then they will always converge to mutual defection of $(D,D)$ eventually. For the rest of cases, there exist infinite number of equilibrium points in-between the above two extreme cases, all of which are not stable, which means that the learning dynamic will never converge to those equilibrium points.

Next we turn to analyze the dynamics of SA-IGA playing the Symmetric Coordination game. The general form of a Coordination game is shown in Table \ref{tab:codgame}. From the table, we can see that the Coordination game is asymmetric if any of the following conditions are met: $R \ne r$, $P \ne p$, $T\ne t$ or $S \ne s$. we analyze a simplified game first, i.e., the Symmetric Coordination game, the general circumstance of coordination game will be analyzed later. Similar to the analysis of Theorem \ref{pdgdynamics}, we have,

\begin{table}[h]
\centering
\begin{tabular}{ccccc}
\hline
\rule{0pt}{12pt}
\multirow{2}{0.6 in}{1's payoff \\ 2's payoff} &  & \multicolumn{2}{c}{Agent 2's actions} \\
\cline{3-4}
\rule{0pt}{12pt}
 & &  1 &  2 \\
\hline
\multirow{2}{0.5 in}{\\Agent 1's \\ actions}\\
 &  1 & R/r & S/t \\
\cline{2-4}
\rule{0pt}{12pt}
 &  2 & T/s & P/p \\
\cline{2-4}
\hline
\end{tabular}
\caption{The General Form of a Coordination Game (where $R>T\wedge P>S$ and $r>t\wedge p>s$)} \label{tab:codgame}
\end{table}

\begin{corollary}
\label{Scoodynamics}
The dynamics of SA-IGA algorithm under a symmetric coordination game have two types of stable equilibrium points:
\begin{enumerate}
\item $\left(1,1,w_{1}^{*},w_{2}^{*}\right)$, with $\frac{{T - S}}{2}w_1^*>T-R$;
\item $\left(0,0,w_{1}^{*},w_{2}^{*}\right)$, with $\frac{{T - S}}{2}w_i^*<P-S$;
\end{enumerate}
where $u=R+P-S-T$. 
\begin{proof}
: The proof is the same with Theorem \ref{pdgdynamics}, thus we omit it here.
\end{proof}
\end{corollary}

Intuitively, for a Symmetric Coordination game, from Corollary \ref{Scoodynamics}, there are two types of stable equilibrium if players playing SA-IGA policy, which means players will eventually converging to action $(1,1)$ or $(0,0)$, i.e., the Nash equilibriums of the Symmetric Coordination game. Besides, because the final convergence of the algorithm depends on the combined effect of $p$ and $w$, we cannot give a theoretical conclusion about the condition under which the algorithm will converge to the social optimal for a symmetric Coordination game. In fact, experimental simulations in the following section show that the SA-IGA has a higher probability converging to social optimal.

Now we turn to consider the asymmetric case. As we mentioned before, SA-IGA under an asymmetric game may have nonlinear dynamics when $u_1 \ne u_2$, which has caused great difficulties for theoretical analysis. For this reason, we only analyze the general Coordination game which is a typical asymmetric game.

\begin{theorem}
\label{coodynamics}
The dynamics of SA-IGA algorithm under a general coordination game have three types of equilibrium points:
\begin{enumerate}
\item $\left(0,0,w_{1}^{*},w_{2}^{*}\right)$, with $w_{1}^{*}=1\wedge w_{2}^{*}=0$ when $P>p>t$; $w_{1}^{*}=0\wedge w_{2}^{*}=1$ when $T<P<p$; and $\left(\frac{t-S}{2}{{w}_{1}^{*}}<P-S\right)\wedge \left(\frac{T-s}{2}{{w}_{2}^{*}}< p-s\right)$ when $P=p$;
\item $\left(1,1,w_{1}^{*},w_{2}^{*}\right)$, with $w_{1}^{*}=1\wedge w_{2}^{*}=0$ when $R>r>s$; $w_{1}^{*}=0\wedge w_{2}^{*}=1$ when $T<R<r$; and $\left(\frac{T-s}{2}{{w}_{1}^{*}}< R-T\right)\wedge \left(\frac{S-t}{2}{{w}_{2}^{*}}< r-t\right)$ when $R=r$;
\item $\left(p_{1}^{*},p_{2}^{*},w_{1}^{*},w_{2}^{*}\right)$, others.
\end{enumerate}

The first and second types of equilibrium points are stable, while the last non-boundary equilibrium points is not. 
\begin{proof}
: Following the system of differential equations in Equations (\ref{eq3.7differential}), we can express the dynamics of SA-IGA in coordination game as follows:
\begin{equation}
\begin{split}
&\dot{p}_{1}=\left(u_{1}+\frac{u_{2}-u_{1}}{2}w_{1}\right )p_{2}+\frac{d_{2}-c_{1}}{2}w_{1}+c_{1}\\
&\dot{p}_{2}=\left(u_{2}+\frac{u_{1}-u_{2}}{2}w_{2}\right )p_{1}+\frac{d_{1}-c_{2}}{2}w_{2}+c_{2}\\
&\dot{w}_{1}=\varepsilon \cdot \left[\left (u_{1}-u_{2}\right)p_{1}p_{1}+\left(c_{1}-c_{2}\right)p_{1}+\left(d_{2}-d_{1}\right)p_{2}+e_1\right]\\
&\dot{w}_{2}=-\dot{w}_{1}\\
\end{split}
\end{equation}
where $\varepsilon=\frac{\eta_{w}}{\eta_{p}} >0$,$u_{1}=R+P-S-T>0$, $u_{2}=r+p-s-t>0$, $c_{1}=S-P$, $c_{2}=s-p$, $d_{1}=T-P$, $d_{2}=t-p$, and $e_1=P-p$.
We can see that the dynamic of coordination game is nonlinear when $u_{1}\ne u_{2}$. We start with proving the last type of equilibrium points first:

If there exit a equilibrium $eq=\left(p_{1}^{*},p_{2}^{*},w_{1}^{*},w_{2}^{*}\right)^{T}\in\left(0,1\right)^{4}$, then there have $\dot{p}_{i}\left(eq\right)=0$ and $\dot{w}_{i}\left(eq\right)=0$, $i\in \{1,2\}$. By linearizing the unconstrained update differential equations into the form of $\dot{x}=Ax+B$ in point $eq=\left(p_{1}^{*},p_{2}^{*},w_{1}^{*},w_{2}^{*}\right)^{T}$, we have

\[
A=\begin{bmatrix}
0&u_{1}^{*}&a_{13}&0 \\
u_{2}^{*}&0&0&a_{24} \\
-\varepsilon a_{13}&\varepsilon a_{24}&0&0\\
\varepsilon a_{13}&-\varepsilon a_{24}&0&0
\end{bmatrix}
\]

where $u_{1}^{*}={{u}_{1}}+\frac{{{u}_{2}}-{{u}_{1}}}{2}w_{1}^{*}$ and $u_{2}^{*}={{u}_{2}}+\frac{{{u}_{1}}-{{u}_{2}}}{2}w_{2}^{*}$, 
The parameters $a_{ij}$ are represented as functions of $p_{1}^{*},p_{2}^{*},w_{1}^{*}$ and $w_{2}^{*}$.
Without loss of generality, we set ${u}_{1}\ge{u}_{2}$. Because of ${u}_{1}\ge {u}_{2}>0$, and $w_{1}^{*},w_{2}^{*}\in [0,1]$, we have $u_{1}^{*}\in [ \frac{{{u}_{1}}+{{u}_{2}}}{2}\text{,}{{u}_{1}}]$ and $u_{2}^{*}\in [{{u}_{2}}\text{,}\frac{{{u}_{1}}+{{u}_{2}}}{2}]$, which means $u_{1}^{*}>u_{2}^{*}>0$.

After calculating matrix $A$'s eigenvalue in Matlab, we have an eigenvalue $\lambda_{1}=0$, an eigenvalue $\lambda_{2}$ with its real part  $Re\left(\lambda_{2}\right)>0$, an eigenvalue $\lambda_{3}$ with $Re\left(\lambda_{3}\right)<0$ and an eigenvalue $\lambda_{4}$ close to $0$. Since there exists an eigenvalue $\lambda>0$, the equilibrium $eq$ is not stable\cite{shilnikov2001methods}.

Next we turn to prove the first type of equilibrium. In this case, we need to put the projection function back since we are dealing with boundary cases.

For the case $P>p>t$, we have $V_{i}^{\mathrm{idv}}\left(eq\right)>V_i^{\mathrm{soc}}\left(eq\right)$, thus $\dot{w}_{r}\left(eq\right)>0$ and $\dot{w}_{2}\left(eq\right)<0$, which means ${w}_{1}$ and ${w}_{2}$ will keep ${w}_{1}=1$ and ${w}_{2}=0$. Because $\dot{p}_{1}\left(eq\right)=\frac{t-p+S-P}{2}<0$ and $\dot{p}_{2}\left(eq\right)=s-p<0$, then ${p}_{r}$ and ${p}_{c}$ will keep ${p}_{r}=0$ and ${p}_{c}=0$. According to the continuity theorem of differential equations \cite{Coddington1955Theory}, $\left(0,0,1,0\right)$ is a stable equilibrium. The case $p>P>T$ can be proved similarly, which is omitted here.

For the case $P=p$, we have $V_{i}^{\mathrm{idv}} = V_i^{\mathrm{soc}}$, then $\dot{w}_{1}\left(eq\right)=-\dot{w}_{2}\left(eq\right)=\varepsilon \left(V_{1}^{idv} -V_i^{soc}\right)=0$. Because $\left(\frac{T-s}{2}{{w}_{2}^{*}}< p-s\right)$, we have $\dot{p}_{1}=\frac{T-s}{2}{{w}_{2}^{*}}+ s-p< 0$. Because $\left(\frac{t-S}{2}{{w}_{1}^{*}}< P-S\right)$, we have $\dot{p}_{2}=\frac{t-S}{2}{{w}_{2}^{*}}+ S-P< 0$. According to the continuity theorem of differential equations, $\left(0,0,w_{1}^{*},w_{2}^{*}\right)$ is a stable equilibrium.
The stability of the second type of equilibrium points can be proved similarly, which is omitted here.
\end{proof}
\end{theorem}

From Theorem \ref{coodynamics}, we find that conclusions of Corollary \ref{Scoodynamics} is a special case of Theorem \ref{coodynamics}. Note that it can be verified by drawn the symmetry conditions into Theorem \ref{coodynamics}.

\section{A Practical Algorithm}\label{sec-algorithm}

In SA-IGA, each agent needs to know the policy of others and the payoff function, which are usually not available before a repeated game starts. Based on the idea of SA-IGA, we relax the above assumptions and propose a practical multiagent learning algorithm called Socially-Aware Policy Gradient Ascent (SA-PGA). The overall flow of SA-PGA is shown in Algorithm \ref{alg:SA-PGA}. In SA-PGA, each agent only needs to observe the payoffs of both agents by the end of each round.

\begin{algorithm}
\caption{SA-PGA for player $i$}
\label{alg:SA-PGA}
\begin{algorithmic}[1]
\STATE  Let $\alpha_{\pi}$,$\alpha_{w}\in \left (0,1\right )$ and $\beta \in \left ( 0,1 \right )$ be learning rates.
\STATE Initialize,\\
 $Q_{i}^{\mathrm{idv}}\left (a\right )\leftarrow 0$, $Q_{i}^{\mathrm{soc}}\left (a\right )\leftarrow 0$, $Q_{i}\left (a\right )\leftarrow 0$, \\
 $\pi_i(a)\leftarrow \frac{1}{|A_i|}$, $w_i\leftarrow w_0$.
\REPEAT
\STATE Same as PHC in Step 4 of Table \ref{alg:PHC}.
\STATE Observing reward $r$ and the average of all agents' current rewards $r_{\mathrm{all}}$,\\
  $Q_{i}^{\mathrm{idv}}\left (a \right ) \leftarrow \left (1-\beta \right )Q_{i}^{\mathrm{idv}}\left (a \right )+\beta r$,\\
  $Q_i^{\mathrm{soc}}\left ( a \right ) \leftarrow \left (1-\beta \right )Q_i^{\mathrm{soc}}\left (a \right )+\beta r_{\mathrm{all}}$,\\
  $Q_i\left (a \right ) \leftarrow \left (1-w_i\right )Q_i^{\mathrm{idv}}\left (a \right )+w_{i}Q_i^{\mathrm{soc}}\left (a \right )$.\\
\STATE Update $\pi_i$ according to gradient ascent strategy, Same as PHC in Step 6 of Table \ref{alg:PHC}.\\
\STATE Update $w_i$,\\
  $V_i^{\mathrm{idv}} = \sum\nolimits_{a \in {A_i}} {{\pi _i}(a)Q_i^{\mathrm{idv}}(a)} $ .\\
  $V_i^{\mathrm{soc}} = \sum\nolimits_{a \in {A_i}} {{\pi _i}(a)Q_i^{\mathrm{soc}}(a)} $ .\\
  $w_i \leftarrow \Pi_{\left[0,1\right]} [w_i + \alpha_{w}\left(V_i^{\mathrm{idv}}- V_i^{\mathrm{soc}} \right)]$ .\\
\UNTIL{the repeated game ends}
\end{algorithmic}
\end{algorithm}


In SA-IGA, we know that agent $i$'s policy (the probability of selecting each action) is updated based on the partial derivative of the expected value $V_i$, while the social attitude $w$ is adjusted according to the relative value of $V_i^{idv}$ and $V_i^{soc}$. Here in SA-PGA, we first estimate the value of $V_i^{idv}$ and $V_i^{soc}$ using Q-values, which are updated based on the immediate payoffs received during repeated interactions. Specifically, each agent $i$ keeps a record of the Q-value of each action for both its own and the average of all agents ($Q_i^{idv}$ and $Q_i^{soc}$) (Step 5). Both Q-values are updated following Q-learning update rules accordingly by the end of each round (Step 5). Then the overall Q-value of each agent is calculated as the weighted average of $Q_i^{idv}$ and $Q_i^{soc}$ weighted by its social attitude $w$ (Step 5). The policy update strategy is the same as the Table \ref{alg:PHC} in Step 6. Finally, the social attitude of agent $i$ is updated in Step 7. The value of $V_i^{\mathrm{idv}}$ and $V_i^{\mathrm{soc}}$ are estimated based on its current policy and Q-values. The updating direction of $w_i$ is estimated as the difference between $V_i^{idv}$ and $V_i^{soc}$. Note that a SA-PGA player in each interaction needs only to know its own reward and the average reward of all agents. Knowing the average reward of a group is a reasonable assumption in many realistic scenarios, such as elections and voting. 

\section{Experimental Evaluation}\label{experiment}
This section is divided into three parts. Subsection \ref{exp.1} compare SA-IGA and SA-PGA with simulation in different types of two-agents, two-actions, general-sum games. Subsection \ref{exp.2} presents the experimental results for the 2x2 benchmark games, specifically, performance of converging to the social optimal outcomes and against selfish agents. Subsection \ref{exp.3} presents the experimental results for games with multiple agents, i.e. public good game\cite{Andreoni1998Partners}.
\subsection{Simulation comparison of SA-IGA and SA-PGA}\label{exp.1}
We start the performance evaluation with analyzing the learning performance of SA-PGA under two-player two-action repeated games. In general a two-player two-action game can be classified into three categories\cite{tuyls2006evolutionary}:
\begin{enumerate}
  \item $\exists i \in \{1,2\}$, $(r_{i}^{11}-r_{i}^{21})(r_{i}^{12}-r_{i}^{22})>0$. In this case, each player has a dominant strategy and thus the game only has one pure strategy NE.
  \item $\forall i \in \{1,2\}$, $(r_{i}^{11}-r_{i}^{21})(r_{i}^{12}-r_{i}^{22})<0$  and $(r_{1}^{11}-r_{1}^{21})(r_{2}^{21}-r_{2}^{22})>0$. In this case, there are two pure strategy NEs and one mixed strategy NE.
  \item $\forall i \in \{1,2\}$, $(r_{i}^{11}-r_{i}^{21})(r_{i}^{12}-r_{i}^{22})<0$ and $(r_{1}^{11}-r_{1}^{21})(r_{2}^{21}-r_{2}^{22})<0$. In this case, there only exists one one mixed strategy NE.
\end{enumerate}
where $r_{i}^{jk}$ is the payoff of player $i$ when player $i$ takes action $j$ while its opponent $-i$ takes action $k$. We select one representative game for each category for illustration.

\subsubsection{Category 1}
For category 1, we consider the PD game as shown in Table \ref{tab:pdg1}. In this game, both players have one dominant strategy $D$, and $(D, D)$ is the only pure strategy NE, while there also exists one socially optimal outcome $(C, C)$ under which both players can obtain higher payoffs.

\begin{figure*}[h]
 \centering
\subfigure[SA-PGA in PD game]{\label{pdsimulation}\includegraphics[width=55mm]{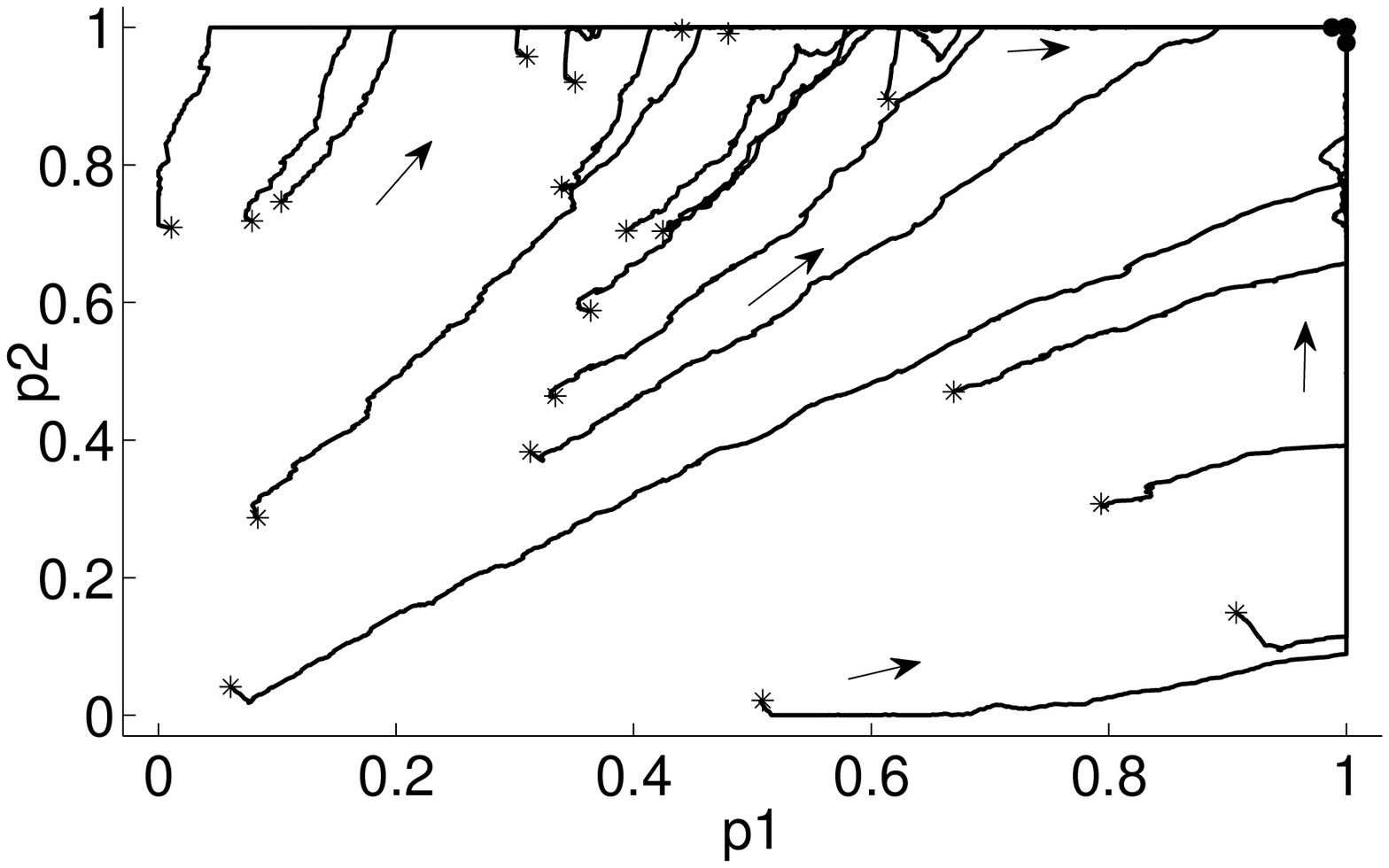}}
\subfigure[SA-IGA in PD game]{\label{pdthoeretical}\includegraphics[width=55mm]{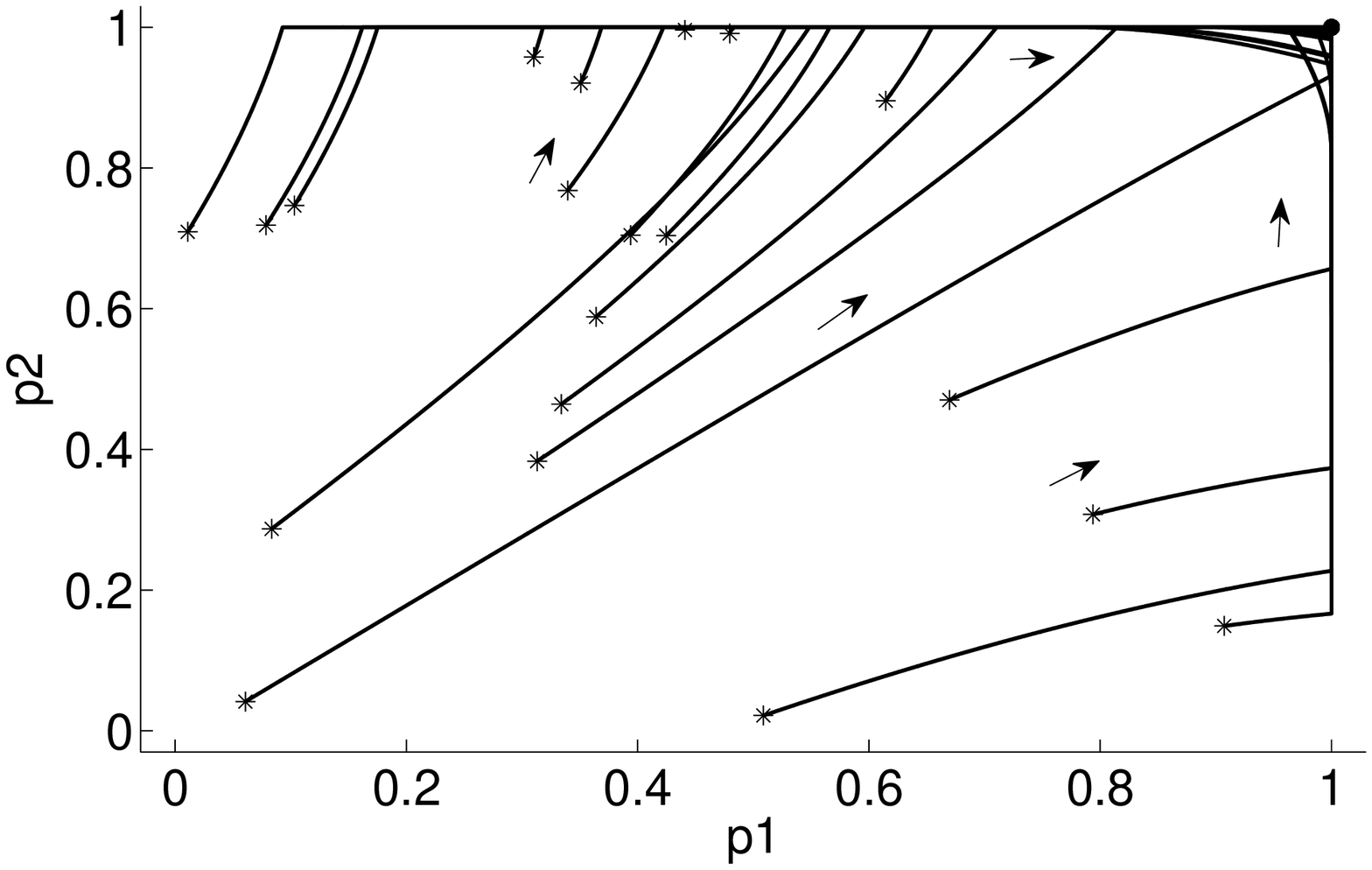}}
\caption{The Learning Dynamics of SA-IGA  and SA-PGA in PD game ( ${{w}_{1}}(0)={{w}_{2}}(0)=0.85$, $\alpha_{\pi}=\alpha_{w}=0.001$, $\beta=0.8$)}
\label{fig:games1}
\end{figure*}

Figure 1(a)
show the learning dynamics of the practical SA-PGA algorithm playing the PD game. The x-axis $p1$ represents player 1's probability of playing action $C$ and the y-axis $p2$ represents player 2's probability of playing action $C$. We randomly selected 20 initial policy points as the starting point for the SA-PGA agents. We can observe that the SA-PGA agents are able to converge to the mutual cooperation equilibrium point starting from different initial policies.

Figure 1(b)
illustrates the learning dynamics predicted by the theoretical SA-IGA approach. Similar to the setting in Figure \ref{pdsimulation}, the same set of initial policy points are selected and we plot all the learning curves accordingly. We can see that for each starting policy point, the learning dynamics predicted from the theoretical SA-IGA is well consistent with the learning curves from simulation. This indicates that we can better understand and predict the dynamics of SA-PGA algorithm using its corresponding theoretical SA-IGA model.

\subsubsection{Category 2}
For category 2, we consider the CG game as shown in Table \ref{tab:coordination}. In this game, there exist two pure strategy Nash equilibria (C, C) and (D, D), and both of them are also socially optimal.

\begin{table}[h]
\centering
\begin{tabular}{ccccc}
\hline
\rule{0pt}{12pt}
\multirow{2}{0.6 in}{1's payoff \\ 2's payoff} &  & \multicolumn{2}{c}{Agent 2's actions} \\
\cline{3-4}
\rule{0pt}{12pt}
 & &  C &  D \\
\hline
\multirow{2}{0.5 in}{\\Agent 1's \\ actions}\\
 & C & 3/4 & 0/0 \\
\cline{2-4}
\rule{0pt}{12pt}
 & D & 0/0 & 4/3 \\
\cline{2-4}
\hline
\end{tabular}
\caption{Coordination game (Category 2)} \label{tab:coordination}
\end{table}

\begin{figure*}[h]
 \centering
\subfigure[SA-PGA in CG]{\label{coosimulation}\includegraphics[width=55mm]{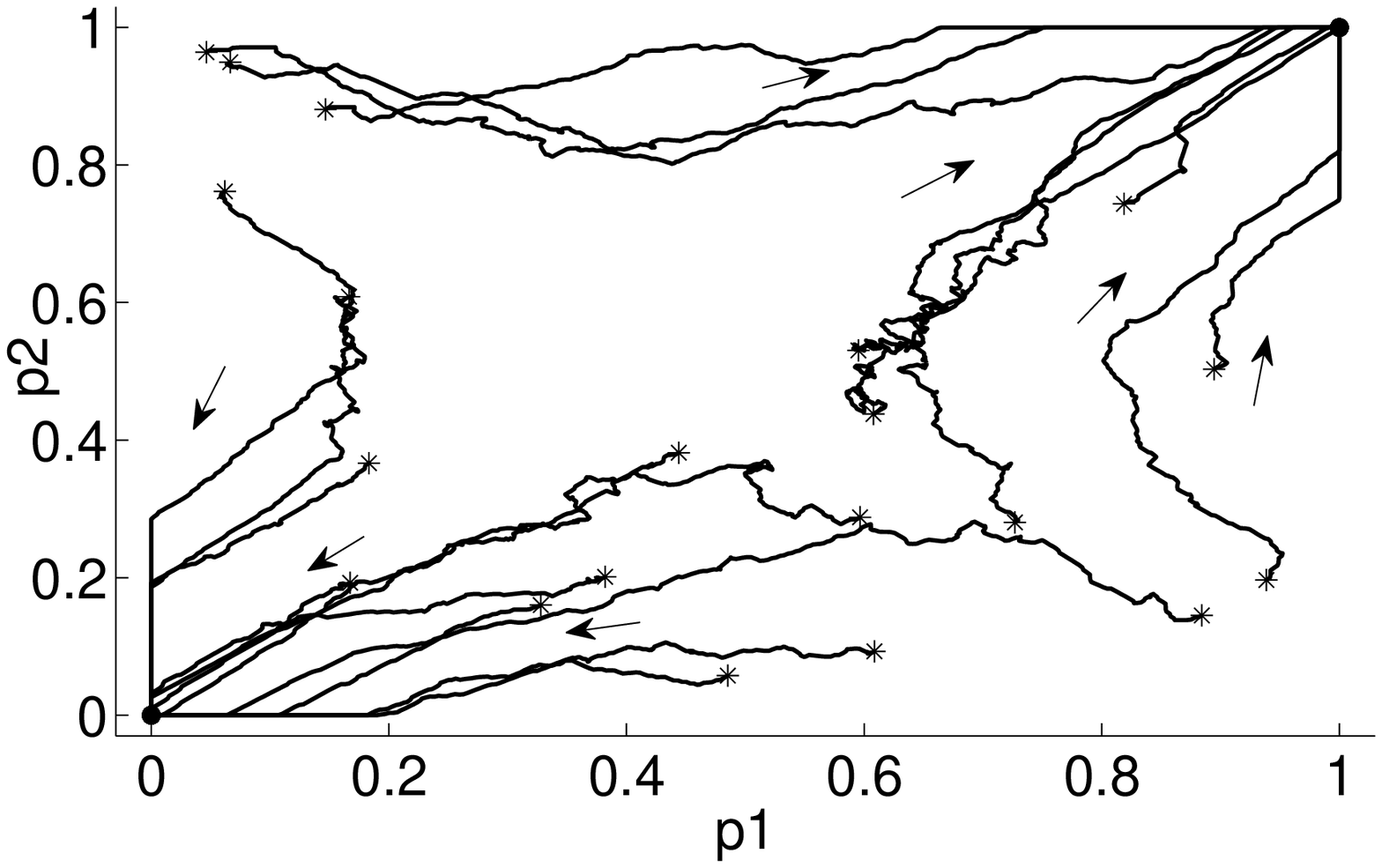}}
\subfigure[SA-IGA in CG]{\label{cootheoretic}\includegraphics[width=55mm]{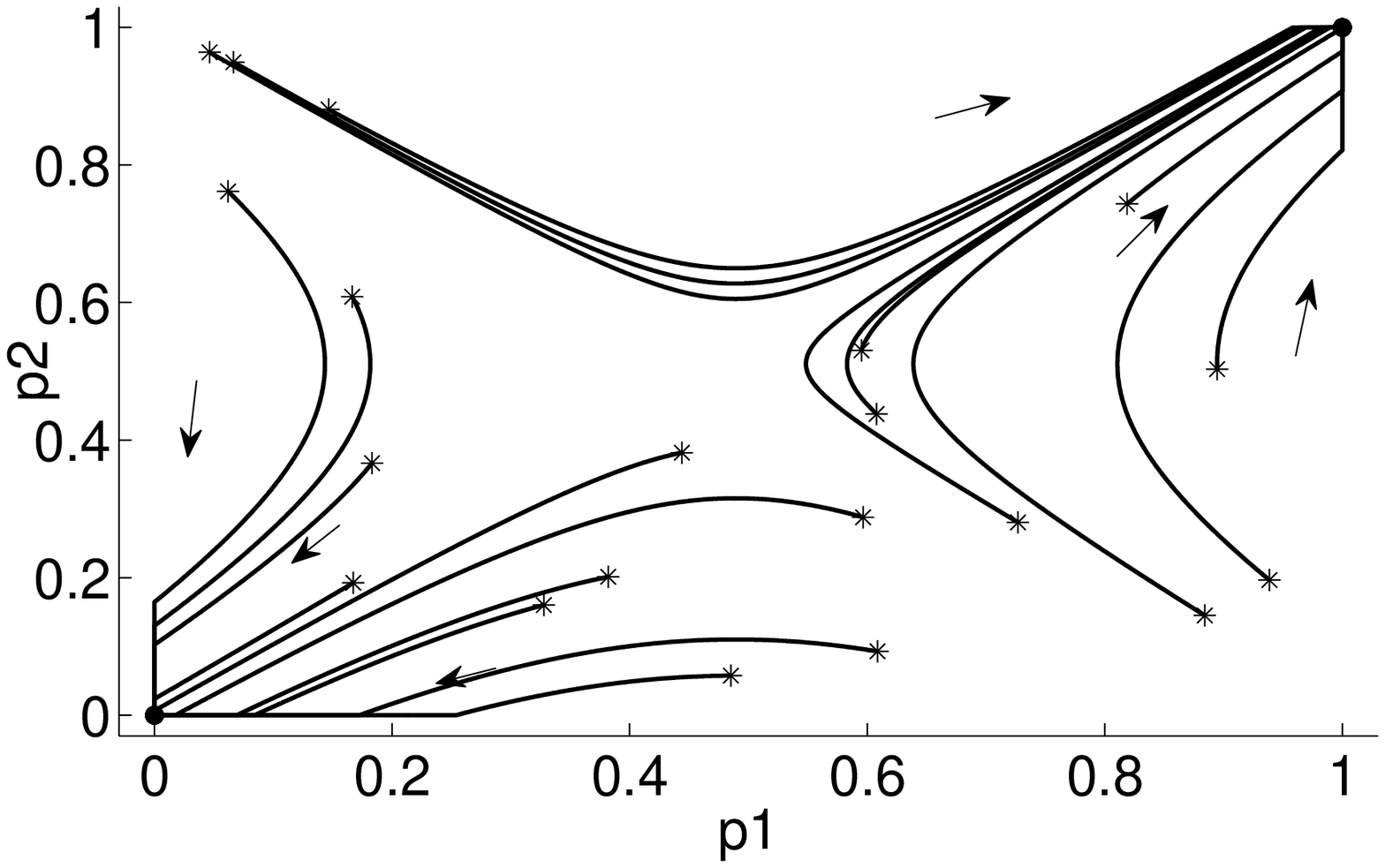}}
\caption{The Learning Dynamics of SA-IGA  and SA-PGA in coordination game ( ${{w}_{1}}(0)={{w}_{2}}(0)=0.85$, $\alpha_{\pi}=\alpha_{w}=0.001$, $\beta=0.8$)}
\label{fig:games2} 
\end{figure*}

Figure 2(a)
illustrates the learning dynamics of the practical SA-PGA algorithm playing a CG game. The x-axis $p1$ represents player 1's probability of playing action $C$ and the y-axis $p2$ represents player 2's probability of playing action $C$. Similar to the case of PD game, 20 initial policy points are randomly selected as the starting points. We can see that the SA-PGA agents can converge to either of the aforementioned two equilibrium points depending on the initial policies they start with.

Figure 2(b)
shows the learning dynamics predicted by the theoretical SA-IGA approach. Similar to the setting in Figure \ref{coosimulation}, we adopt the same set of 20 initial policy points for comparison purpose. All the learning curves starting from these 20 policy points are drawn accordingly. We can observe that for each starting policy point, the learning dynamics predicted from the theoretical SA-IGA is well consistent with the learning curves obtained from simulation. Therefore, the theoretical model can facilitate better understanding and predicting the dynamics of SA-PGA algorithm.

\subsubsection{Category 3}

The game we use in Category 3 is shown in Table \ref{tab:gtype3}. In this game, there only exists one mixed strategy Nash equilibrium, while the pure strategy outcome $(C, D)$ is socially optimal.

\begin{table}[!h]
\centering
\begin{tabular}{ccccc}
\hline
\rule{0pt}{12pt}
\multirow{2}{0.6 in}{1's payoff \\ 2's payoff} &  & \multicolumn{2}{c}{Agent 2's actions} \\
\cline{3-4}
\rule{0pt}{12pt}
 & &  C &  D \\
\hline
\multirow{2}{0.5 in}{\\Agent 1's \\ actions}\\
 & C & 3/2 & 4/4 \\
\cline{2-4}
\rule{0pt}{12pt}
 & D & 1/3 & 5/1 \\
\cline{2-4}
\hline
\end{tabular}
\caption{An example game of Category 3} \label{tab:gtype3}
\end{table}

\begin{figure*}[h]
 \centering
\subfigure[SA-PGA for the game with one mix NE]{\label{thirdsimulation}\includegraphics[width=55mm]{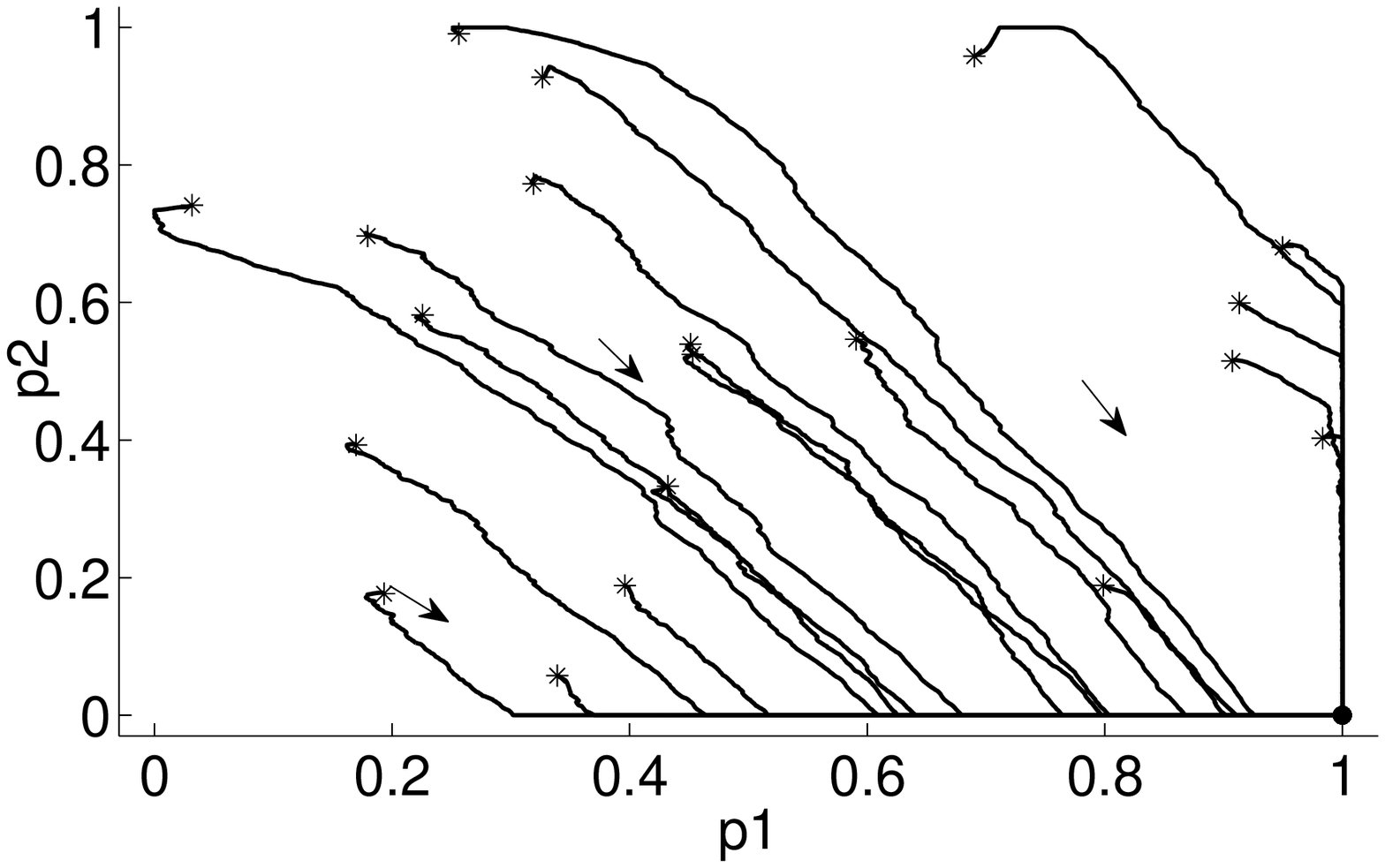}}
\subfigure[SA-IGA for the game with one mix NE]{\label{thridthoeretical}\includegraphics[width=55mm]{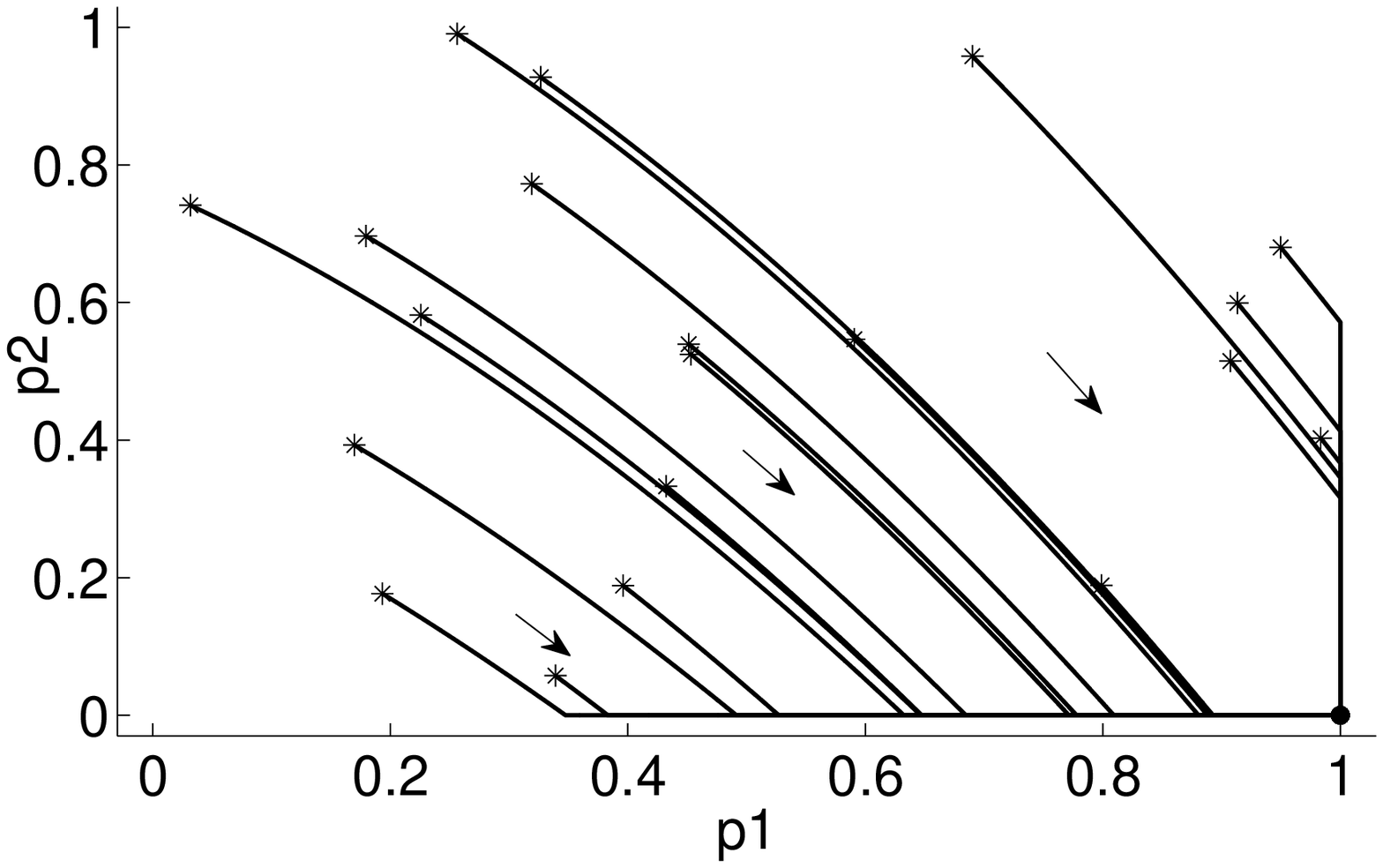}}
\caption{The Learning Dynamics of SA-IGA and SA-PGA in game with one mix NE ( ${{w}_{1}}(0)={{w}_{2}}(0)=0.85$, $\alpha_{\pi}=\alpha_{w}=0.001$, $\beta=0.8$)}
\label{fig:games3} 
\end{figure*}

Figure 3(a)
illustrates the learning dynamics of the practical SA-PGA algorithm playing the game in Table \ref{tab:gtype3}. The x-axis $p1$ and y-axis $p2$ represent player 1's probability of playing action $C$ and player 2's probability of playing action $C$ respectively. Similar to the previous cases, 20 initial policy points are randomly selected as the starting points. From Figure \ref{thirdsimulation}, we can see that the SA-PGA agents can always converge to the socially optimal outcome $(C, D)$ no matter where the initial policies start with.

Figure 3(b)
presents the learning dynamics of agents predicted by the theoretical SA-IGA approach. Similar to the setting in Figure \ref{thirdsimulation}, we adopt the same set of 20 initial policy points for comparison purpose, and the corresponding learning curves are drawn accordingly. From Figure \ref{thridthoeretical}, we can observe that for each starting policy point, the theoretical SA-IGA model can well predict the simulation results of SA-PGA algorithm. Therefore, better understanding and insights of the dynamics of SA-PGA algorithm can be obtained through analyzing its corresponding theoretical model.

\subsection{Performance in $2\times2$ General-sum Games}\label{exp.2}
In this subsection we turn to evaluate the performance of SA-PGA in two-agents, two-actions, general-sum games. First we implement two previous representative learning algorithms for comparison: CJAL \cite{CJAL} and WoLF-PHC\cite{WOLF-PHC}. We compare their performance based on the following two criteria: utilitarian social welfare and Nash social welfare, which reflect the system-level efficiency of different learning strategies in terms of the total payoffs received for the agents. Then we evaluate the ability of SA-PGA against selfish opponents with the same three representative games used in previous sections.
\subsubsection{Comparison of SA-PGA with CJAL and WoLF-PHC}
we evaluate the performance of SA-PGA with CJAL \cite{CJAL} and WoLF-PHC \cite{WOLF-PHC} in two-player's repeated games under self-play. CJAL is selected since this algorithm is specifically designed to enable agents to achieve mutual cooperation (i.e., maximizing social welfare) instead of inefficient NE for games like prisoner's dilemma. WoLF-PHC is selected as one representative NE-oriented algorithm for baseline comparison purpose. For all previous strategies the same parameter settings used in their original papers are adopted.

\begin{table}[!h]
\centering
\begin{tabular}{  p{3.1cm} | p{3.0cm} | p{3.0cm} }
      \hline
      & Utilitarian Social Welfare & Nash Product \\ \hline
      SA-PGA (our strategy) (${{w}_{r}}(0)={{w}_{c}}(0)=0.85$) &$7.241\pm0.003$&$12.706\pm0.015$\\
      CJAL \cite{CJAL}	&$6.504\pm0.032$&$10.887\pm0.114$ \\
      WoLF-IGA \cite{WOLF-PHC}& $6.536\pm0.004$ & $10.943\pm0.145$\\
    \hline
    \end{tabular}
\label{comparisonresult11}
\caption{Performance comparison with CJAL and WoLF-PHC}
\end{table}

We use all possible structurally distinct two-player, two-action conflict games as a testbed for SA-PGA. In each game, each player ranks the four possible outcomes from 1 to 4. We use the rank of an outcome as the payoff to that player for any outcome. We perform the evaluation under 100 randomly generated games with strict ordinal payoffs. We perform 10,000 interactions for each run and the results are averaged over 20 runs for each game.

We compare their performance based on the the following two criteria: utilitarian social welfare (USW) and Nash social welfare (NSW). Utilitarian social welfare is the sum of the payoffs obtained by the two players in their converged state, while Nash social welfare is the product of the payoffs obtained by two players in their converged state. Formally, ${\rm{USW = }}{V_1}{\rm{ + }}{V_2}$ and $NSW = {V_1}{V_2}$, where $V_1$ and $V_2$ are payoffs obtained by the two players in their converged state, averaged over 100 randomly generated games. Both criteria reflect the system-level efficiency of different learning strategies in terms of the total payoffs received for the agents. Besides, Nash social welfare also partially reflects the fairness in terms of how equal the agents' payoffs are. The overall comparison results are summarized in Table 6. We can see that SA-PGA outperforms the previous CJAL strategy and WoLF-PHC strategy under both criteria. 
\subsubsection{Performance Against Selfish Agents}

\begin{figure}[!h]
\centering
\includegraphics[width=75mm,height=40mm]{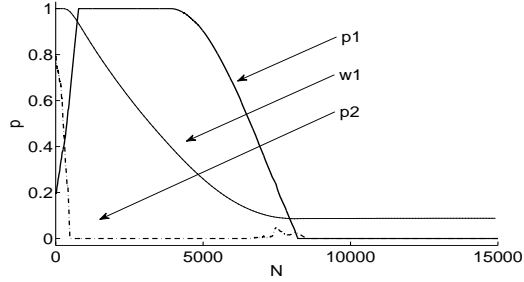}
\caption{SA-PGA against a selfish agent for in PD game(${{w}_{r}}(0)=1$, ${{p}_{r}}(0)=0.2$ and ${{p}_{c}}(0)=0.8$)}
\label{againstselfishpd}
\end{figure}

\begin{figure}[!h]
\centering
\includegraphics[width=75mm,height=40mm]{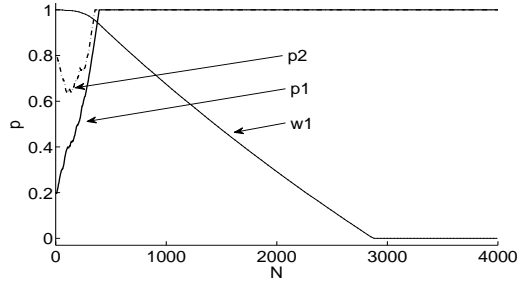}
\caption{SA-PGA against a selfish agent for in coordination game(${{w}_{r}}(0)=1$, ${{p}_{r}}(0)=0.2$ and ${{p}_{c}}(0)=0.8$)}
\label{againstselfishcoor}
\end{figure}

\begin{figure}[!h]
\centering
\includegraphics[width=75mm,height=40mm]{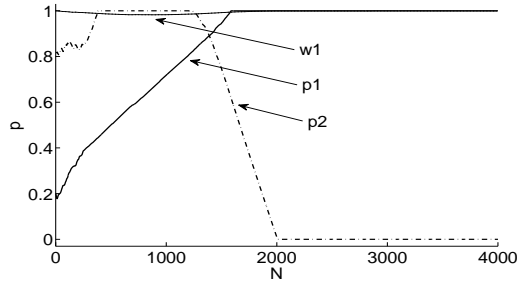}
\caption{SA-PGA against a selfish agent for the game with only one mix NE(${{w}_{r}}(0)=1$, ${{p}_{r}}(0)=0.2$ and ${{p}_{c}}(0)=0.8$)}
\label{againstselfishthird}
\end{figure}

If a learning agent is facing selfish agents that attempt to exploit others, one reasonable choice for an effective algorithm is to learn a Nash equilibrium. In this section, we evaluate the ability of SA-PGA against selfish opponents. We adopt the same three representative games used in previous sections as the testbed and the results are given in Figure \ref{againstselfishpd}, \ref{againstselfishcoor} and \ref{againstselfishthird} respectively. We can observe that for the PD and coordination games, the SA-PGA agent can successfully achieve the corresponding NE solution. This property is desirable since it prevents the SA-PGA agent from being taken advantage by selfish opponents. The results also show how the socially-aware degree $w$ of SA-PGA agent changes, which varies depending on the game structure. For PD and coordination game, a SA-PGA agent eventually behaves as a purely individually rational entity and one pure strategy NE is eventually converged to. In contrast, for the third type of game (Table \ref{tab:gtype3}), a SA-PGA agent behaves as a purely socially rational agent and cooperate with the selfish agent towards the socially optimal outcome $(C, D)$ without fully exploiting the opponent. This indicates the cleverness of SA-PGA since higher individual payoff can be achieved under the outcome (C, D) than pursuing Nash equilibrium $(C, C)$.

\subsection{Performance in games with multiple agents}\label{exp.3}

We use Public Goods Game (PGG)\cite{Andreoni1998Partners} to further evaluate the performance of SA-PGA in multiple agent cases. PGG is an extended version of the PD game in multiagent environment, which has attracted increasing attention to study cooperative behavior and, in particular, deviations from the ¡°rational¡± equilibrium \cite{Shivshankar2015An,Wei2013Cooperation}. In a typical public goods experiment a group of players is endowed with one dollar each. The players then have the opportunity to invest their money into a common pool, knowing that the total amount will be doubled and split equally among all players, irrespective of their contributions. If everybody invests their money, they end up with two dollars. However, each player faces the temptation to free-ride on others' contributions by withholding the money because each invested dollar yields only a return of 50 cents. If everybody adopts this ¡°rational¡± strategy, no one would increase the initial capital and forego the public good. The payoffs for cooperators $R_C$ and defectors $R_D$ in a group of $N$ interacting individuals are then given by,


\[
{R_D} = \frac{r N_C c }{N}, {R_C} = {R_D} - c
\]

where $r$ denotes the multiplication factor of the public good, $N_C$ the number of cooperators in the group and $c$ the cost of the cooperative contributions, i.e. each agent's investment in the public good. 
From the definition, the defect action, i.e., the action of not contributing to the public, is the dominate strategy because $R_D>R_C$. The Nash equilibrium of all PGG players is that everyone chooses to defect, while the social optimal outcomes strategy of PGG is that everyone contributes the the public. We evaluate the performance of SA-PGA in PGG repeated games with three players under three circumstance: 1) games with three SA-PGA players, and 2) games with two SA-PGA player and one selfish opponent, and 3) games with one SA-PGA player and two selfish opponents. Without lose of generality, all players' initial policies $p(0)$ of each game are settled to $0.5$. Other parameters such as $r$ and $c$ in the three experiments are exactly the same, $r=2$, $c=2$.

\begin{figure}[h!]
\centering
\includegraphics[width=75mm,height=40mm]{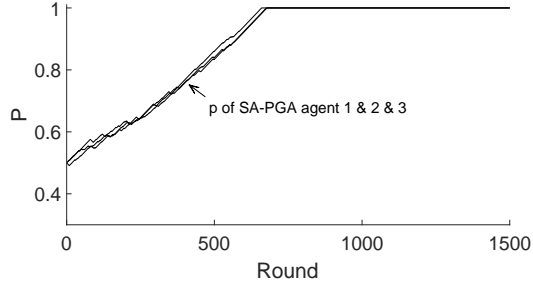}
\caption{The Learning Dynamics of SA-PGA in PGG with three SA-PGA players ($\alpha_{\pi}=\alpha_{w}=0.001$, $\beta=0.8$, $p_i(0)=0.5$ and $w_i(0)=0.85$)}
\label{fig:pgg1}
\end{figure}

\begin{figure}[h!]
\centering
\includegraphics[width=75mm,height=40mm]{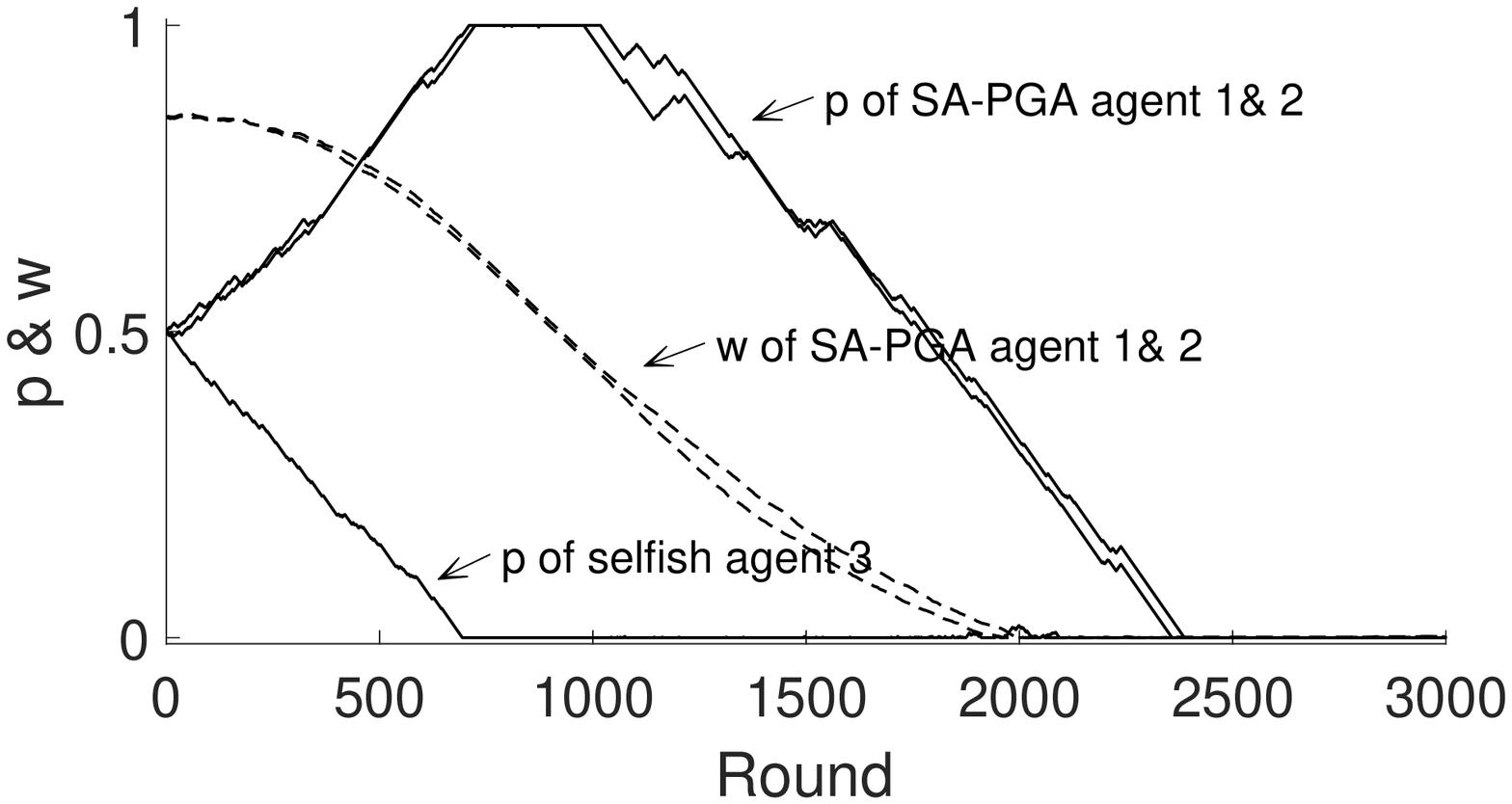}
\caption{The Learning Dynamics of SA-PGA in PGG with two SA-PGA players and one selfish opponents ($\alpha_{\pi}=\alpha_{w}=0.001$, $\beta=0.8$, $p_i(0)=0.5$ and $w_1(0)=w_2(0)=0.85$)}
\label{fig:pgg2}
\end{figure}

\begin{figure}[h!]
\centering
\includegraphics[width=75mm,height=40mm]{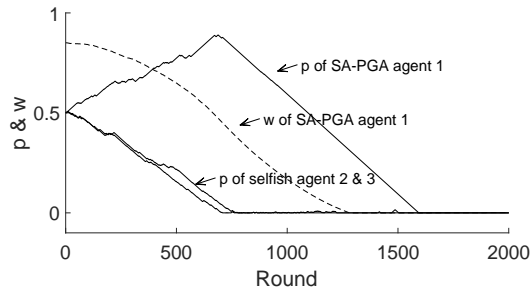}
\caption{The Learning Dynamics of SA-PGA in PGG with one SA-PGA players and two selfish opponents ($\alpha_{\pi}=\alpha_{w}=0.001$, $\beta=0.8$, $p_i(0)=0.5$ and $w_1(0)=0.85$)}
\label{fig:pgg3}
\end{figure}

Figure \ref{fig:pgg1} shows the learning dynamics of PGG games with three SA-PGA players. The y-axis $p$ represents the probability of playing action $C$, i.e. the cooperate action, while the x-axis $t$ is the timeline. Each line in Figure \ref{fig:pgg1} shows the learning dynamic of one player's strategy. We can observe that the SA-PGA agents are able to converge to the mutual cooperation equilibrium point giving the initial value of $w(0)$ large enough (here we set $w = 0.85$).

Figure \ref{fig:pgg2} shows the learning dynamics of PGG games with two SA-PGA players and one selfish opponent, while Figure \ref{fig:pgg3} shows the learning dynamics of PGG games with one SA-PGA player and two selfish opponents. The y-axis $p\& w$ represents the probability of strategy $p$ and the socially-aware degree $w$. The solid lines are learning dynamics of players' strategies, and dotted lines are learning dynamics of SA-PGA players' socially-aware degrees. From Figure \ref{fig:pgg2} and \ref{fig:pgg3}, we can observe that agents initially tends to cooperate with others and later realizes that the other agents are not cooperating, thus converging to the pure strategy $D$ eventually behaves as a purely individually rational entity. This property is desirable since it prevents the SA-PGA agent from being taken advantage by selfish opponents.



\section{Conclusion}\label{conclusion}
In this paper, we proposed a novel way of incorporating social awareness into traditional gradient-ascent algorithm to facilitate reaching mutually beneficial solutions (e.g., (C, C) in PD game). We first present a theoretical gradient-ascent based policy updating approach (SA-IGA) and analyzed its learning dynamics using dynamical system theory. For PD games, we show that mutual cooperation (C,C) is stable equilibrium point as long as both agents are strongly socially-aware. For Coordination games, either of the Nash equilibria (C,C) and (D,D) can be a stable equilibrium point depending on the agents' socially-aware degrees. Following that, we proposed a practical learning algorithm SA-PGA relaxing the impractical assumptions of SA-IGA. Experimental results show that a SA-PGA agent can achieve higher social welfare than previous algorithms under self-play and also is robust against individually rational opponents. As future work, more testbed scenarios (e.g., population of agents)  will be applied to further evaluate the performance of SA-PGA. Another interesting direction is to investigate how to further improve the convergence rate of SA-PGA.


\bibliographystyle{spmpsci}      
\bibliography{Jaamas}

\end{document}